\pdfoutput=1
\documentclass[11pt]{article}

\usepackage[preprint]{acl}

\usepackage{times}
\usepackage{latexsym}
\usepackage{algorithm}
\usepackage{algorithmic}

\usepackage[T1]{fontenc}
\usepackage[utf8]{inputenc}

\usepackage{microtype}

\usepackage{inconsolata}

\usepackage{graphicx}
\usepackage{tikz}

\usepackage{amsmath}
\usepackage{booktabs}
\usepackage{threeparttable}
\usepackage{multirow}
\usepackage{multicol}
\usepackage{makecell}

\usepackage{amsmath,amsfonts,bm}

\def\eqref#1{equation~\ref{#1}}
\def\1{\bm{1}}

\def\vw{{\bm{w}}}

\DeclareMathAlphabet{\mathsfit}{\encodingdefault}{\sfdefault}{m}{sl}
\SetMathAlphabet{\mathsfit}{bold}{\encodingdefault}{\sfdefault}{bx}{n}

\def\gW{{\mathcal{W}}}

\newcommand{\modelparams}{\boldsymbol{\theta}}
\newcommand{\proxyparams}{\boldsymbol{\phi}}
\newcommand{\bestmodel}{\boldsymbol{\theta}^{\star}}
\newcommand{\weights}{\vw}

\newcommand{\bestweights}{\vw^{\star}}
\newcommand{\getweights}{\text{Get\_Weights}}

\title{Rethinking Data Mixture for Large Language Models: A Comprehensive Survey and New Perspectives}

\author{
Yajiao Liu$^1$, Congliang Chen$^1$, Junchi Yang$^{1,2}$, Ruoyu Sun\thanks{Corresponding Author.}$^{1,2}$ \vspace{0.2cm} \\
$^1$The Chinese University of Hong Kong, Shenzhen \\
$^2$Shenzhen Research Institute of Big Data \\
\texttt{\{yajiaoliu, congliangchen\}@link.cuhk.edu.cn} \\
\texttt{\{yangjunchi, sunruoyu\}@.cuhk.edu.cn} \vspace{0.2cm} \\
}

\begin{document}
\maketitle
\vspace{-0.3cm}
\begin{abstract}
% Collecting data from various domains can increase the diversity of training data, further improving the generalization capability of large language models on downstream tasks.
Training large language models with data collected from various domains can improve their performance on downstream tasks.
However, given a fixed training budget, the sampling proportions of these different domains significantly impact the model's performance.
How can we determine the domain weights across different data domains to train the best-performing model within constrained computational resources?
In this paper, we provide a comprehensive overview of existing data mixture methods.
First, we propose a fine-grained categorization of existing methods, extending beyond the previous offline and online classification.
Offline methods are further grouped into heuristic-based, algorithm-based, and function fitting-based methods. 
For online methods, we categorize them into three groups—online min-max optimization, online mixing law, and other approaches—by drawing connections with the optimization frameworks underlying offline methods.
Second, we summarize the problem formulations, representative algorithms for each subtype of offline and online methods, and clarify the relationships and distinctions among them.
Finally, we discuss the advantages and disadvantages of each method and highlight key challenges in the field of data mixture. 
\end{abstract}

\vspace{-0.3cm}
\section{Introduction}
\vspace{-0.2cm}

Recently, many works have focused on improving the generalization capability of large language models (LLMs) on downstream tasks by increasing the diversity of training data.
This is typically achieved by collecting data from various domains or different sources (e.g., CommonCrawl, Wikipedia, and Github), either during the pre-training phase~\cite{radford2019language, gao2020pile800gbdatasetdiverse}, or instruction fine-tuning phase~\cite{wei2022finetunedlanguagemodelszeroshot, chung2022scalinginstructionfinetunedlanguagemodels}.
However, given a fixed training budget, the sampling proportions of these different domains or data sources significantly impact the model's performance~\cite{xie2023doremi, albalak2023efficientonlinedatamixing, shen2024slimpajamadcunderstandingdatacombinations}.
This raises an important question: how can we determine the \textbf{domain weights} (the sampling probabilities for each domain) across different data domains to train the best-performing model within constrained computational resources?
\textbf{Data mixture}, also known as domain reweighting, is an emerging research area dedicated to addressing this challenge. It focuses on optimizing the domain weights to sample different domain data, thereby improving the effectiveness and efficiency of language model training.

Many early works rely on heuristically chosen or manually assigned domain weights for training language models~\cite{radford2019language, gao2020pile800gbdatasetdiverse}. 
However, subsequent studies~\cite{xie2023doremi, fan2023doge, albalak2023efficientonlinedatamixing} argue that such approaches are likely suboptimal and propose more principled methods to determine the domain weights automatically.
\citet{albalak2024surveydataselectionlanguage} provide an overview of this rapidly developing field, categorizing existing data mixture methods into two main types: \textbf{offline methods}, which determine domain weights separately from the training of the target model, and \textbf{online methods}, which dynamically adjust domain weights during the target model’s training process.
As the cost of training LLMs continues to rise with the increasing scale and diverse sources of data, data mixture has become increasingly important.
Meanwhile, many new studies—beyond the scope of~\citet{albalak2024surveydataselectionlanguage}—have gradually emerged.
Accordingly, we believe it is timely to provide a more comprehensive and updated review for the community.
A detailed comparison between our survey and~\citet{albalak2024surveydataselectionlanguage} is provided in Appendix~\ref{survey-diff}.

In this paper, we propose a more fine-grained categorization of existing data mixture methods, building on the framework proposed by~\citet{albalak2024surveydataselectionlanguage}.
Offline methods assume that, for a given corpus composed of multiple domains, there exist optimal static domain weights that lead to the best-performing model.
Based on this assumption, they typically estimate these weights using proxy models, which are typically smaller than the target model to reduce the additional training costs, and then apply the estimated weights to train the target model, as described in Algorithm~\ref{offline}.
We divide offline methods into three subcategories: heuristic-based methods, algorithm-based methods, and function fitting-based methods.
\textbf{Heuristic-based methods}, such as uniform sampling and proportional sampling, determine domain weights based on simple heuristics. Although they do not rely on proxy models, we treat them as degenerate cases of offline methods for completeness in our categorization.
\textbf{Algorithm-based methods} apply specific algorithms to determine the optimal domain weights with proxy models.
Specifically, we realize that data mixture should be regarded as a subfield of Multi-Domain Learning (MDL), which is often closely associated with Multi-Task Learning (MTL)~\footnote{One major difference between Data Mixture and the previous methods like MDL and MTL is that, in Data Mixture, all domains share the same set of neural network parameters.}. 
Some methods proposed for MTL can also address MDL, and vice versa~\cite{yang2015unifiedperspectivemultidomainmultitask}.
Therefore, we systematically analyze the connections between existing algorithm-based methods and two prominent optimization frameworks widely used in MTL: \textbf{min-max optimization} and \textbf{bi-level optimization}, aiming to offer a deeper understanding and valuable insights for optimization strategies in data mixture.
\textbf{Function fitting-based methods} empirically learn a mapping from domain weights to model performance by conducting multiple training runs on proxy models using different domain weights. Once this function—also referred to as a \textbf{mixing law}—is learned, it can be used to infer the optimal domain weights for training the target model.

% \vspace{-0.3cm}
\begin{algorithm}[t]
\caption{Framework of Offline methods}
\label{offline}
\begin{algorithmic}[1]
    \STATE {\bf Input:} Proxy model $\proxyparams$, target model $\modelparams$, training data $\mathcal D_{\text{train}}$, validation data $\mathcal D_{\text{val}}$ (which can be \text{optional});
    \STATE {\bf Initialization:} $\weights^0$;
    \STATE $\bestweights = \getweights(\weights^0, \proxyparams, \mathcal D_{\text{train}}, \mathcal D_{\text{val}})$;
    \STATE $\bestmodel = \arg\min_{\modelparams} \sum_{i=1}^k w^{\star}_i \ell_{\text{train},i}(\modelparams)$;
    \STATE {\bf Output:} $\bestweights$ and $\bestmodel$.
\end{algorithmic}
\end{algorithm}

% \vspace{-0.4cm}

Online methods jointly optimize the target model and the domain weights during training, as illustrated in Algorithm~\ref{online}.
During the target model training, the domain weights are dynamically optimized to effectively and efficiently steer the target model toward optimal performance.
Most online methods can be regarded as online variants of existing offline approaches.
For example, certain offline methods can be transformed into their online counterparts by replacing the proxy models used for optimizing domain weights with the target model itself, thereby enabling joint optimization of both domain weights and the target model during training.
From this perspective, we categorize existing online methods into three groups: \textbf{online min-max optimization} methods, \textbf{online mixing law} methods, and other approaches.

% \vspace{-0.3cm}
\begin{algorithm}[t]
\caption{Framework of Online methods}
\label{online}
\begin{algorithmic}[1]
\STATE {\bf Input:} Target model $\modelparams$, training data $\mathcal D_{\text{train}}$, validation data $\mathcal D_{\text{val}}$ (\text{optional}), the total number of training steps $T$, the update interval $t_{\text{update}}$;
\STATE {\bf Initialization:} The target model $\modelparams^0$, and $\weights^0$;
    \FOR {$t = 1,2, \cdots, T$}
    \STATE $\modelparams^{t} = \text{Update\_}\modelparams (\weights^{t-1}, \modelparams^{t-1}, \mathcal D_{\text{train}})$;
    \IF{$t$ mod $t_{\text{update}} = 0$}
        \STATE $\weights^t = \text{Update\_}\weights(\weights^{t-1}, \modelparams^{t}, \mathcal D_{\text{train}}, \mathcal D_{\text{val}})$; 
    \ENDIF
    \ENDFOR
    \STATE {\bf Output:} $\modelparams^{T}$.
\end{algorithmic}
\end{algorithm}
% \vspace{-0.2cm}

Our contributions can be summarized as follows:
1. We propose a fine-grained categorization of existing data mixture methods, extending beyond the previous offline and online classification. 
Specifically, we divide offline methods into heuristic-based, algorithm-based, and function fitting-based methods. For online methods, we categorize them into three groups—online min-max optimization, online mixing law, and other approaches—by drawing connections with the optimization frameworks underlying offline methods.
2. For each subtype of offline and online methods, we summarize their problem formulations, representative algorithms, and clarify the relationships and distinctions among them, aiming to offer a deeper understanding and valuable insights for optimization strategies in data mixture to the community.
3. Finally, we discuss the advantages and disadvantages of each method and highlight key challenges and future directions in this field.

\vspace{-0.2cm}
\section{Notations}
\vspace{-0.2cm}

\label{notations}

Given $k$ data domains, each domain $ i \in [k] = \{ 1, 2, \cdots, k \} $ is associated with a training dataset $D_{\text{train}, i}$ and a validation dataset $D_{\text{val}, i}$. 
We represent a composition of the training data $\mathcal D_{\text{train}} = \left\{D_{\text{train}, 1}, \ldots, D_{\text{train}, k}\right\}$ using domain weights $\weights = [w_1, w_2, \cdots, w_k]^T$, where $w_i$ denotes the weight assigned to domain $i$ and $\sum_{i=1}^k w_i = 1, w _i \geq 0$ (i.e., $\weights \in \mathbb{R}_{\geq 0}^k$ lies within a probabilistic simplex $\Delta^k$).
Our ultimate goal is to obtain the best-performing model $\bestmodel$, where $\modelparams$ is used to parameterize the target model, by optimizing the domain weights $\weights$ across the training data $\mathcal {D}_{\text{train}}$. 
We use $\proxyparams$ to represent the parameters of the proxy model.

The best-performing model refers to the one that achieves the highest overall performance across various downstream tasks. 
However, during the training of a model $\modelparams$, we typically optimize a loss function $\ell(\modelparams): \mathbb{R}^d \rightarrow \mathbb{R}{\geq 0}$ as the surrogate objective for its downstream task performance, which is often the negative log-likelihood in the context of LLMs. 
For example, in the pretraining setting, the loss can be written as $\ell(x;\modelparams)=-\log p_{\modelparams}(x)=-\frac{1}{|x|}\sum_{z=1}^{|x|} \log p_{\modelparams}(x_{z} \mid x_{1:z-1})$, where $|x|$ denotes the number of tokens in a given example $x$.
There are typically two settings for computing this loss: on the training data or the validation data.
We denote the average loss of model $\modelparams$ on each training domain $D_{\text{train}, i}$, $i \in [k]$ as $\ell_{\text{train},i}(\modelparams)$ (i.e., $\ell_{\text{train},i}(\modelparams) = \ell(D_{\text{train},i}; \modelparams)) = \frac{1}{|D_{\text{train},i}|} \sum_{x \in D_{\text{train},i}} \ell(x;\modelparams))$).
For validation data, we consider two cases: \textbf{in-domain} validation data $\mathcal D_{\text{val}}^{\text{in}} = \{D_{\text{val}, 1}, \cdots, D_{\text{val}, k}\}$, which correspond to the same domains as the training data $\mathcal D_{\text{train}}$, and \textbf{out-of-domain} validation data ${\mathcal D}_{\text{val}}^{\text{ood}} = \{\hat  D_{\text{val}, 1}, \cdots, \hat D_{\text{val}, m}\}$, consisting of $m$ domains completely unrelated to $\mathcal D_{\text{train}}$~\footnote{In fact, the \textbf{out-of-domain} validation data ${\mathcal D}_{\text{val}}^{\text{ood}}$ could contain several domains related to the training data $\mathcal D_{\text{train}}$. For simplicity, we assume that all domains in ${\mathcal D}_{\text{val}}^{\text{ood}}$ are completely unrelated to $\mathcal D_{\text{train}}$.}.
We denote the average loss of $\modelparams$ on $D_{\text{val}, j}, j \in [k]$ as $\ell_{\text{val}, j}(\modelparams) = \ell(D_{\text{val},j}; \modelparams)$, on $\hat D_{\text{val}, j}, j \in [m]$ as $\hat \ell_{\text{val}, j}(\modelparams) =  \ell(\hat D_{\text{val},j}; \modelparams)$.

We provide a detailed explanation of \textbf{static weights} and \textbf{dynamic weights} in Appendix~\ref{static-dynamic} to clarify how different methods leverage them to approximate the best-performing model $\modelparams^{\star}$.

\vspace{-0.2 cm}
\section{Offline Methods}
\label{offline-methods}

Offline methods assume the existence of optimal static domain weights $\bestweights$ for a given corpus composed of multiple domains.
Based on the strategies used to find $\bestweights$ (i.e., line 3 of Algorithm~\ref{offline}), we categorize offline methods into three groups:
heuristic methods (Section~\ref{Heuristic}), algorithm-based methods (Section~\ref{Algorithm}), and function fitting-based methods (Section~\ref{learning-methods}).

\begin{table*}[t]
\vspace{-0.7cm}
  \small
  \centering
  % \resizebox{\linewidth}{!}{
  \begin{tabular}{ l | l | l | l }
    \toprule
    \multicolumn{2}{c|}{Category} & Method & Training phase(s) \\
    \midrule
    \multirow{11}{*}{Offline} & Heuristic & Uniform / Proportional sampling  & PT /  CPT / SFT \\
    \cmidrule(r){2-4}

    & \makecell[l]{Algorithm-based \\ (Min-max)} &  DoReMi~\cite{xie2023doremi}  & PT \\
    \cmidrule(r){2-4}
    
    & \multirow{2}{*}{\makecell[l]{Algorithm-based \\ (Bi-level)}} & DoGE~\cite{fan2023doge}  & PT  \\
    & & ScaleBiO~\cite{pan2024scalebio} & SFT \\
    \cmidrule(r){2-4}
    
    & \multirow{6}{*}{Function fitting-based} & DML~\cite{ye2024datamixinglaws} & PT \\
    & & BiMix~\cite{ge2024bimixbivariatedatamixing} & PT \\
    & & D-CPT~\cite{que2024dcptlawdomainspecificcontinual} & CPT \\
    & & AutoScale~\cite{kang2025autoscalescaleawaredatamixing} & PT \\
    & & RegMix~\cite{liu2024regmix} & PT \\
    & & MDE~\cite{belenki2025optimizingpretrainingdatamixtures} & PT \\
    \midrule
    
    \multirow{8}{*}{Online} & \multirow{3}{*}{Online min-max} & MFTCoder~\cite{liu2023mftcoderboostingcodellms} & SFT  \\
    & & ShearedLLaMa~\cite{xia2024shearedllamaacceleratinglanguage} & CPT  \\
    & & Velocitune~\cite{luo2024velocitunevelocitybaseddynamicdomain} & CPT \\
    \cmidrule(r){2-4}
    & \multirow{2}{*}{Online mixing laws} & Skill-It~\cite{chen2023skillitdatadrivenskillsframework} & CPT / SFT \\
    & & Aoili~\cite{chen2024aioliunifiedoptimizationframework} & PT \\
    \cmidrule(r){2-4}
    & \multirow{3}{*}{Others} & ODM~\cite{albalak2023efficientonlinedatamixing} & PT \\
    & & ADO~\cite{jiang2024adaptivedataoptimizationdynamic} & PT \\
    \bottomrule
  \end{tabular}
  % }
  \vspace{-0.2cm}
  \caption{Overview of existing data mixture methods discussed in this paper.  
  In the last column, we report the specific training setting for each method, as described in the corresponding original paper.
  SFT denotes Supervised Fine-Tuning for instruction tuning; PT refers to Pretraining, and CPT stands for Continued Pretraining. } 
  \label{Overview}
\vspace{-0.4cm}
\end{table*}

\vspace{-0.1cm}
\subsection{Heuristic-based Methods}
\label{Heuristic}

Heuristic-based methods determine $\bestweights$ using predefined heuristics.
For example, uniform sampling assigns equal weights to all $k$ domains, i.e., $\bestweights = [\frac{1}{k}, \dots, \frac{1}{k}]$.
Since these methods can be implemented by omitting line 3 in Framework~\ref{offline}, we consider them as degenerate cases of offline methods, even though they do not rely on proxy models.  
Further details are provided in Appendix~\ref{app:Heuristic}.

\vspace{-0.1cm}
\subsection{Algorithm-based Methods}
\label{Algorithm}
\vspace{-0.1cm}

Next, we systematically survey existing algorithm-based methods for data mixture under two optimization frameworks: \textbf{min-max optimization} and \textbf{bi-level optimization}.
The definition of the best-performing model $\modelparams^{\star}$ can vary depending on the specific setting--when the model's loss is computed solely using training data or incorporating validation data.
This distinction leads to different optimization algorithms being employed to find $\bestweights$.

\vspace{-0.1cm}
\subsubsection{Min-Max Optimization}
\label{Offline_train}

Given all $k$ domains, we could determine the $\boldsymbol{w}^{\star}$ by focusing on whether the proxy model $\proxyparams$ is well-optimized on each domain of the training data $\mathcal D_{\text{train}}$ without considering any validation data.
One surrogate objective to achieve this is to ensure that the worst-case performance across domains remains satisfactory, which can be formulated as the following min-max problem:
\begin{equation}
\setlength\abovedisplayskip{3pt}
\label{eq:weighted_loss}
\small
\begin{aligned}
\min _{\proxyparams} \max_{\boldsymbol{w} \in \Delta^k} \sum_{i=1}^k w_i f_i(\proxyparams),
\end{aligned}
\setlength\belowdisplayskip{3pt}
\end{equation}
where $f_i(\proxyparams)$ is often defined based on $\ell_{\text{train},i}(\proxyparams)$, the average training loss of $\proxyparams$ on the $i$-th domain $D_{\text{train}, i}$, for $i \in [k]$.

\vspace{-0.1cm}
\paragraph{Min-max optimization.} Equation~\ref{eq:weighted_loss} represents a nonconvex-concave min-max optimization problem~\cite{pmlr-v119-lin20a, Razaviyayn_2020}.
The well-known gradient descent ascent (GDA) algorithm~\cite{daskalakis2018limitpointsoptimisticgradient, pmlr-v119-lin20a}, a simple extension of gradient descent (GD) to the min-max setting, is one of the most popular algorithms for solving this problem, capable of finding a reasonably good solution for the proxy model $\proxyparams$ and domain weights $\weights$.

\citet{xie2023doremi} propose Domain Reweighting with Minimax Optimization (DoReMi) to achieve the same goal in the LLMs pretraining setting with $f_i(\proxyparams) = \ell_{\text{train},i}(\proxyparams) - \ell_{\text{train},i}(\proxyparams_{\text{ref}}) $, where $\proxyparams_{\text{ref}}$ represents a pre-trained reference model, and its loss $\ell_{\text{train},i}(\proxyparams_{\text{ref}}))$ serves as the baseline for the loss $\ell_{\text{train}, i}(\proxyparams)$.
The objective is to minimize the worst-case excess loss across domains, which is the gap in loss between $\proxyparams$ and $\proxyparams_{\text{ref}}$.
They adopt the Group DRO~\cite{sagawa2020distributionallyrobustneuralnetworks} to solve this problem, which can be viewed as a variant of GDA.
Since $\weights$ lies within the probabilistic simplex $\Delta^k$, at each step $t$, $\weights^t$ is updated using exponential gradient ascent (i.e., mirror ascent).
Then, the average weights are returned as the optimized domain weights (i.e., $\bestweights=\frac{1}{T} \sum_{i=1}^T \weights^t$).
Finally, they use $\bestweights$ to resample the training data $\mathcal D_{\text{train}}$ for training a larger target model $\modelparams$.
There are other variants of GDA, such as Optimistic GDA~\cite{daskalakis2018traininggansoptimism} and Smoothed GDA~\cite{zhang2022singleloopsmoothedgradientdescentascent}, which can also be used to solve Equation~\ref{eq:weighted_loss}.

\vspace{-0.2cm}
\subsubsection{Bi-Level Optimization}
\vspace{-0.1cm}
\label{Offline_val}

Ensuring that the model is well-optimized on each domain in the training data $\mathcal{D}_{\text{train}}$ implicitly assumes that good optimization leads to good generalization.
However, we can explicitly enhance the model's generalization ability by finding $\bestweights$ that maximize its performance on the validation data $\mathcal D_{\text{val}}$.
In this case, whether evaluated using in-domain or out-of-domain validation loss, this objective can be formulated as the following standard bi-level optimization problem~\cite{franceschi2017forwardreversegradientbasedhyperparameter, lorraine2019optimizingmillionshyperparametersimplicit}:
\begin{equation}
\small
\label{eq:bi-level}
\begin{aligned}
& \boldsymbol{w}^{\star} \in \underset{\boldsymbol{w} \in \Delta^k}{\arg \min } \ell_{\text{val}} \left(\proxyparams^{\star}(\boldsymbol{w})\right) \\
& \quad \text { s.t. } \proxyparams^{\star}(\boldsymbol{w}) \in \underset{\proxyparams}{\arg \min } \sum_{i=1}^k w_{i} \ell_{\text{train}, i}(\proxyparams),
\end{aligned}
\setlength\belowdisplayskip{3pt}
\end{equation}
where $\ell_{\text{val}} \left(\proxyparams^{\star}(\boldsymbol{w})\right)$ can be $\frac{1}{k} \sum_{j=1}^{k} \ell_{\text{val},j} \left(\proxyparams^{\star}(\boldsymbol{w})\right)$
or
$\frac{1}{m} \sum_{j=1}^{m} \hat \ell_{\text{val},j} \left(\proxyparams^{\star}(\boldsymbol{w})\right)\ $ for considering the in-domain validation data $\mathcal D_{\text{val}}^{\text{in}}$ or the out-of-domain validation data ${\mathcal D}_{\text{val}}^{\text{ood}}$.

\begin{table*}[t]
\vspace{-0.7 cm}
  \small
  \centering
  \resizebox{\linewidth}{!}{
  \begin{tabular}{ l | c | c | l}
    \toprule
    Method  & Mixing law $f_j, j \in [k]$ & The learning parameters & Objective function (including aggregation function $g$) \\
    \midrule
    DML  & $\ell_{\text{val}, j} \left( \weights \right)= c_j + b_j \exp \left(\sum_{i=1}^k A_{ij} w_i \right)$ & $\boldsymbol{c}, \boldsymbol{b} \in \mathbb{R}^k, \boldsymbol{A} \in \mathbb{R}^{k \times k}$ & $\bestweights = \min_{\weights} \sum_{j=1}^{k} h_j \ell_{\text{val}, j}\left(\weights\right) $ \\
    \cmidrule(r){1-4}

    BiMix  & $\ell_{\text{val},j}\left(w_j,s\right)=\frac{A_j}{w_j^{\alpha_j}}\left(\frac{B_j}{s^{\beta_j}} + C_j\right)$ & $A, B, C, \alpha, \beta \in \mathbb{R}^k$ & $\bestweights = \min_{\weights} \frac{1}{k} \sum_{j=1}^{k} \ell_{\text{val}, j}\left(w_j, s\right) \text{ s.t. } \sum w_j = 1 $ \\
    \cmidrule(r){1-4}
    
    AutoScale & $\ell_{\text{val}, j}\left(w_j, D \right) = \left(D^j_0+ w_j D\right)^{-\gamma_j}+ c_j$ & $D_0, \gamma, c \in \mathbb{R}^k$ & $\bestweights = \min_{\weights} \frac{1}{k} \sum_{j=1}^{k} \ell_{\text{val}, j}\left(w_j, D\right) \text{ s.t. } \sum w_j = 1$\\
    \cmidrule(r){1-4}

    RegMix & $\ell_{\text{val}, j} (\weights) = \mathbf{M}_j(\weights)$  & Regression model $\{\mathbf{M}_j \}_{j=1}^k$ & $\bestweights = \min_{\weights} \ell_{\text{val}, j}\left(\weights\right)$ \\
    \cmidrule(r){1-4}
    MDE  & \makecell{$\ell_{\text{val}, j} (\weights; \text{MDE features}) = \mathbf{M}_j(\weights; L_{\mathrm{MDE}}^1, \ldots, L_{\mathrm{MDE}}^k)$,  \\
    where $L_{\mathrm{MDE}}^j= - \frac{1}{ | \mathcal{D}_{\text{val},j}|} \sum _{x \in \mathcal{D}_{\text{val},j}} \sum_{i=1}^k w_i \log p_{\proxyparams^{\star}_i}(x)$ } & Regression model $\{\mathbf{M}_j \}_{j=1}^k$ & $\bestweights = \min_{\weights} \frac{1}{k} \sum_{j=1}^{k} \ell_{\text{val}, j}\left(\weights, \text{MDE features}\right)$ \\
    \bottomrule
  \end{tabular}
  }
  \vspace{-0.2 cm}
  \caption{The comparison of function fitting-based methods in data mixture.}
  \label{fitting_laws_main}
  \vspace{-0.4 cm}
\end{table*}

\vspace{-0.2cm}
\paragraph{Bi-level optimization.} A bi-level optimization problem involves a two-level hierarchical structure (i.e., outer and inner levels).
The inner-level problem is auxiliary since its solution supports the outer-level problem in finding a better solution.
In equation~\ref{eq:bi-level}, the inner-level problem involves obtaining a good model $\proxyparams^{\star}(\weights)$ trained on a combination of multiple domain data $\mathcal{D}_{\text{train}}$, weighted by domain weights $\weights$.
The outer-level problem aims to find the optimal domain weights $\bestweights$ by evaluating $\proxyparams^{\star}(\boldsymbol{w})$ on the validation data $\mathcal{D}_{\text{val}}$.
There are three primary frameworks for solving bi-level optimization (BLO) problems~\cite{zhang2023introductionbileveloptimizationfoundations}: (1) the Implicit Function (IF)-based approach, (2) the Gradient Unrolling (GU)-based approach, and (3) the Value Function (VF)-based approach.
More details are provided in Appendix~\ref{app:bi-lev}.

\citet{fan2023doge} propose DOmain reweighting with Generalization Estimation (DoGE) to find the optimal domain weights for LLMs pretraining, based on validation data, including $\mathcal D_{\text{val}}^{\text{in}}$ and $\mathcal D_{\text{val}}^{\text{out}}$.
To solve Equation~\ref{eq:bi-level} cheaply, they provide a single-loop Gradient Unrolling (GU)-based approach, which only updates $\proxyparams$ with a single stochastic step in the inner problem.
For the outer-level problem, considering the in-domain validation data $\mathcal D_{\text{val}}^{\text{in}} = \{D_{\text{val}, 1}, \cdots, D_{\text{val}, k}\}$, they update $\weights$ at each step $t$ using a simple first-order update rule: $\weights^t = \weights^{t-1} \odot \exp(\eta^t \gW^{t})$, where $\gW^{t} = [W^t_1, \cdots, W^t_k]$ is a $k$-dimension vector in which each component $W^t_i = \langle\nabla \ell_{\text{train}, i}(\proxyparams^t), \sum_{j \in [k]}\nabla \ell_{\text{val}, j}(\proxyparams^t) \rangle$, $\ell_{\text{val}, j}(\proxyparams^t)$ is the validation loss of $\proxyparams^t$ on $D_{\text{val}, j}$. 
Similar to DoReMi, the final domain weights are averaged over all middle domain weights (i.e., $\bestweights=\frac{1}{T} \sum_{i=1}^T \weights^t$) and then used to resample the training data $\mathcal D_{\text{train}}$ for training the target model $\modelparams$.

\citet{pan2024scalebio} propose ScaleBiO to find the optimal domain weights with proxy model $\proxyparams$ during the supervised instruction fine-tuning (SFT) phase, which is also based on $\mathcal D_{\text{val}}^{\text{in}}$ or $\mathcal D_{\text{val}}^{\text{out}}$. 
They adopt the Fully First-order Stochastic Approximation (F$^2$SA) method introduced by~\citet{kwon2023fullyfirstordermethodstochastic} to solve Equation~\ref{eq:bi-level}, which is a value function (VF)-based bi-level optimization approach. 
By integrating F$^2$SA with the memory-efficient training technique LISA~\cite{pan2024lisalayerwiseimportancesampling}, they successfully scale the bi-level optimization technique to 34-billion-parameter LLMs on eight A40 GPUs.

\vspace{-0.2cm}
\subsection{Function Fitting-based methods}
\label{learning-methods}

Inspired by scaling law~\cite{kaplan2020scalinglawsneurallanguage, hoffmann2022trainingcomputeoptimallargelanguage}, several studies attempt to empirically learn a mapping from domain weights to model performance by conducting multiple training runs on proxy models using different domain weights.
The resulting function enables the prediction of model performance trained with arbitrary domain weights before actual training, thereby facilitating the identification of optimal domain weights $\bestweights$ for a given corpus composed of multiple domains.
We refer to these approaches as function fitting-based methods and the learned functions as \textbf{mixing laws}.

In general, function fitting-based methods evaluate model performance using validation loss. 
For simplicity, we illustrate these methods using in-domain validation loss (the out-of-domain case is analogous), computed over the in-domain validation data $\mathcal{D}_{\text{val}}^{\text{in}} = \{D_{\text{val}, 1}, \cdots, D_{\text{val}, k}\}$.
These methods typically consist of three key components for identifying $\bestweights$: 
\textbf{(1)} fitting a mixing law $f_j$ for each validation domain $j$, $j \in [k]$:
\begin{equation}
\small
\setlength\abovedisplayskip{3pt}
\label{eq:mixing-laws}
\begin{aligned}
f_j & \in \arg\min_{f_j}  \sum_{l=1}^{R} \text{erf}(f_j(\weights_l), \ell_{\text{val}, j}(\proxyparams^{\star}(\weights_l))), \\
 & \text{ where } \proxyparams^{\star}(\weights_l) \in \arg \min _{\proxyparams} \sum_{i=1}^k w_{l, i} \ell_{\text{train},i}(\proxyparams),
\end{aligned}
\setlength\belowdisplayskip{3pt}
\end{equation}
where $\text{erf}(\cdot, \cdot)$ denotes an error function (e.g., mean squared error), and each $f_j$ is fitted using $R$ input–output pairs $\{(\weights_l, \ell_{\text{val},j}(\proxyparams^{\star}(\weights_l))) \}_{l=1}^{R}$.
Each pair is obtained by training a proxy model $\proxyparams$ on $\mathcal{D}_{\text{train}}$ using domain weights $\weights_l$, and then evaluating the resulting model $\proxyparams^{\star}(\weights_l)$ on $D_{\text{val},j}$ to compute its average loss $\ell_{\text{val},j}(\proxyparams^{\star}(\weights_l))$;
\textbf{(2)} defining an aggregation function $g$ to combine all mixing laws into a single objective: $g(f_1(\weights), ..., f_k(\weights))$ to predict model performance for any given $\weights$;
\textbf{(3)} optimizing the objective using appropriate algorithms to identify the optimal domain weights $\bestweights = \min_{\boldsymbol{w} \in \Delta^k} g(f_1(\weights), ..., f_k(\weights))$.

Function fitting-based methods reduce the bi-level optimization problem to a single-level one by fitting mixing laws, which can be solved using gradient-based methods.
We summarize these methods in Table~\ref{fitting_laws_main}. 
For notational convenience, we use $f_j$ and $\ell_{\text{val}, j}$ interchangeably in the following discussion.
\citet{ye2024datamixinglaws} propose the Data Mixing Law (DML), modeling each mixing law as an exponential function of the domain weights $\weights$, as shown in Table~\ref{fitting_laws_main}.
They adopt a weighted sum as the aggregation function $g$ over all mixing laws $f_j$, $j \in [k]$: $g(f_1(\weights), ..., f_k(\weights)) =\sum_{j=1}^{k} h_j \ell_{\text{val}, j}\left(\weights\right)$, where $h_j$ denotes the proportion of validation domain $j$ in the overall validation dataset.
Finally, they perform a grid search over a set of domain weights to identify the one that minimizes the aggregation function $g$ as the optimal domain weights $\bestweights$.
BIMIX~\cite{ge2024bimixbivariatedatamixing} and AutoScale~\cite{kang2025autoscalescaleawaredatamixing} are bivariate data mixing laws that jointly consider domain weights $\weights$ and training data size.
BIMIX represents data size using training steps $s$, whereas AutoScale uses the number of tokens to quantify training data size $D$.
Both methods adopt the average sum as the aggregation function and formulate a constrained optimization problem:
$\bestweights = \min \frac{1}{k} \sum_{j=1}^{k} \ell_{\text{val}, j}(w_j, s \text{ or } D) \text{ s.t. } \sum w_j = 1$. 
BIMIX solves this problem using Lagrange multipliers and numerical methods~\cite{Virtanen_2020}, while AutoScale solves it using projected gradient descent.
We introduce D-CPT law~\cite{que2024dcptlawdomainspecificcontinual} in Appendix~\ref{CD-CPT-law}.

Two recent works learn each mixing law $f_j, j \in [k]$ through regression models. RegMix~\cite{liu2024regmix} adopts a linear regression model or LightGBM—a gradient boosting algorithm that builds an ensemble of decision trees.
Instead of minimizing a weighted or an average sum of all mixing laws, their objective is to minimize the mixing law fitted on a single validation domain, Pile-CC.
\citet{belenki2025optimizingpretrainingdatamixtures} also employ regression models to learn the mixing laws. In addition to domain weights, they propose a Mixture of Data Experts (MDE) to approximate the cross-entropy losses associated with domain weights and use these approximations (i.e., the MDE features) as additional input features.
To obtain the MDE features, they first train a set of data experts $\proxyparams^{\star}_i$, where $i \in [k]$, each on a corresponding training domain $\mathcal{D}_{\text{train}, i}$. 
Then, for each validation domain $j \in [k]$, the MDE feature is computed as:
$L_{\mathrm{MDE}}^j= \frac{1}{ | \mathcal{D}_{\text{val},j}|} \sum _{x \in \mathcal{D}_{\text{val},j}} P_{\mathrm{MDE}}(x, \weights)$, where $P_{\mathrm{MDE}}\left(x, \weights \right):= - \sum_{i=1}^k w_i \log p_{\proxyparams^{\star}_i}(x)$. 
They adopt the average sum as the aggregation function and solve the optimization problem: $\bestweights = \min_{\weights} \frac{1}{k} \sum_{j=1}^{k} \ell_{\text{val}, j}\left(\weights, \text{MDE features}\right)$ using the Vizier framework~\cite{song2024viziergaussianprocessbandit}.

\begin{table*}[t]
  \small
  \vspace{-0.7cm}
  \centering
  % \resizebox{\linewidth}{!}{
  \begin{tabular}{ l | l | c | l}
    \toprule
    Method & The update rules of $\weights^t$ & The value of $t_{\text{update}}$ & The learning progress on each domain $i$\\
    \midrule

    DoReMi$_{\text{online}}$ & $w^t_i = \frac{w^{t-t_{\text{update}}}_i \exp{ (\zeta V^t_i) }}{\sum_{i=1}^k w^{t-t_{\text{update}}}_i \exp{ (\xi V^t_i) }}$ & $t_{\text{update}} = 1$ & $V^t_i = \max \{ \ell_{\text{train}, i} \left(\modelparams^t \right)-\ell_{\text{train}, i}(\modelparams_{\text{ref}}), 0 \}$ \\
    \cmidrule(r){1-4}
    
    ShearedLLaMa &  $w^t_i = \frac{w^{t-t_{\text{update}}}_i \exp{ (V^t_i) }}{ \sum_{i=1}^k w^{t-t_{\text{update}}}_i \exp{ (V^t_i) } }$ & $t_{\text{update}} > 1$ & $V^t_i = \max \{ \ell_{\text{val}, i} \left(\modelparams^t \right)-\ell_{\text{ref}, i}, 0 \}$  \\
    \cmidrule(r){1-4}
    
    Velocitune & $w^t_i = \frac{w^{t-t_{\text{update}}}_i \exp \left( V^t_i \right)}{\sum_{i=1}^k w^{t-\text{update}}_i \exp \left( V^t_i \right)} $ & $t_{\text{update}} > 1$ & $V^t_i = \text{clamp}\{ 0, \frac{\ell_{\text{val}, i} \left(\modelparams^t \right)-\ell_{\text{ref}, i} }{\ell_{\text{init}, i}- \ell_{\text{ref}, i} }, 1 \}$ \\

    \bottomrule
  \end{tabular}
  % }
  \vspace{-0.2cm}
  \caption{
  Overview of three online min-max optimization methods: DoReMi$_\text{online}$, ShearedLLaMa, and Velocitune. 
  These methods share a common update rule for the target model $\modelparams$: $\modelparams^{t} = \modelparams^{t-1} - \eta \sum_{i=1}^k w_i^{t-1} \nabla \ell_{\text{train},i}(\modelparams^{t-1})$, where $\eta$ is the learning rate.
  The dynamic domain weights $\weights^t$ are updated every $t_{\text{update}}$ steps based on $V^t_i$, with $\zeta$ denoting the learning rate for updating $\weights^t$.  Velocitune~\cite{luo2024velocitunevelocitybaseddynamicdomain} refers to $V^t_i$ as the learning velocity of the target model on domain $i$, which we interpret as a measure of learning progress.
  }
  \label{online:min-max}
  \vspace{-0.4 cm}
\end{table*}

\vspace{-0.2cm}
\subsection{Advantages and Disadvantages}

Offline methods explicitly identify the optimal domain weights $\bestweights$ for a corpus composed of multiple domains, which can be beneficial in certain scenarios.
For instance, publicly releasing $\bestweights$ would allow the research community or practitioners to reuse them to train their own models on the same corpus~\cite{mehta2024openelmefficientlanguagemodel}.

However, all offline methods face (1) \textbf{model transfer issue}: ~\citet{albalak2023efficientonlinedatamixing} demonstrates that the optimal domain weights $\bestweights$ may not transfer well across models when adopting new model architectures or different tokenizers.
Algorithm-based methods also face
(2) \textbf{model scale issue}: the optimal weights derived from a small proxy model fail to generalize to larger target models, even if they share the same architecture and tokenizer~\cite{jiang2024adaptivedataoptimizationdynamic, kang2025autoscalescaleawaredatamixing}; 
(3) \textbf{data scale issue}: the optimal weights can shift as the scale of the training data increases~\cite{kang2025autoscalescaleawaredatamixing}.
Function fitting-based methods can potentially address issues (2) and (3) by leveraging scaling laws to extrapolate the optimal weights obtained from small-scale training to larger model sizes and larger training data sizes, though only DML and AutoScale explicitly incorporate this idea into their designs.
Nonetheless, function fitting-based methods also face a scalability issue: as the number of domains increases, the cost of fitting mixing laws grow linearly or even exponentially, making small-scale experiments expensive.

\begin{table*}[t]
  \small
  \centering
  % \resizebox{\linewidth}{!}{
  \begin{tabular}{ l | l | l}
    \toprule
    Method & The update rules of $\weights^t$ &   The learning parameter $A^t_{ij}$ \\
    \cmidrule(r){1-3}
    
    Skill-It &  $w^t_i = \frac{w^{t-t_{\text{update}}}_i \exp{ (\zeta \sum_{j=1}^k A_{ij}^{t-t_{\text{update}}} \ell_{\text{val}, j}(\modelparams^t)) } }{\sum_{i=1}^k w^{t-1}_i \exp{ (\zeta \sum_{j=1}^k A_{ij}^{t-t_{\text{update}}}\ell_{\text{val}, j}(\modelparams^t)) }}$  &  $A_{i j}^t=\ell_{\mathrm{val}, i}^t(\weights)\left(\ell_{\mathrm{val}, i}^{T+1}\left(\mathbf{1}_j\right)-\ell_{\mathrm{val}, i}^1\left(\mathbf{1}_j\right)\right) / \ell_{\mathrm{val}, i}^1\left(\mathbf{1}_j\right)$ \\
    \cmidrule(r){1-3}
    
    Aioli    &  $w^t_i = \frac{w^{t-t_{\text{update}}}_i \exp{ (\zeta \sum_{j=1}^k A_{ij}^{t-t_{\text{update}}}) }}{\sum_{i=1}^k w^{t-1}_i \exp{ (\zeta \sum_{j=1}^k A_{ij}^{t-t_{\text{update}}} ) }}$ & $A_{i j}$ is fitted from $\ell_{\text{val}, j}^{t+1}(\weights) = \ell_{\text{val}, j}^t(\weights) - \sum_{i=1}^k A_{ij}^t w_i^t$ \\

    \bottomrule
  \end{tabular}
  % }
  \vspace{-0.1cm}
  \caption{Overview of two online mixing law methods: Skill-It and Aioli. We provide a simplified version of each method to highlight its main components. Both approaches update the dynamic domain weights $\weights^t$ with $t_{\text{update}} > 1$. $\zeta$ denotes the learning rate for updating $\weights^t$. $T$ denotes the total number of updates to the domain weights $\weights^t$.
  }
  \label{online:mixing laws}
  \vspace{-0.3 cm}
\end{table*}

\vspace{-0.2cm}
\section{Online Methods}
\vspace{-0.1cm}
\label{online-methods}

During target model $\modelparams$ training, online methods employ dynamic domain weights $\weights^t$ to effectively and efficiently steer the target model $\modelparams$ toward optimal performance. 
Following the framework outlined in Algorithm~\ref{online}, most online methods update $\modelparams$ at each training step $t$ (line 4 in Algorithm~\ref{online}) in a similar manner: $\modelparams^{t} = \modelparams^{t-1} - \eta \sum_{i=1}^k w_i^{t-1} \nabla \ell_{\text{train}, i}(\modelparams^{t-1})$, where $\eta$ is the learning rate.
The main difference among these methods lies in how they update the dynamic weights $\weights^t$ (line 6 in Algorithm~\ref{online}).
Based on this difference, we categorize existing online methods into three groups: online min-max optimization, online mixing law, and other methods.

\vspace{-0.2 cm}
\subsection{Online Min-Max Optimization Methods}

Gradient manipulation methods in MTL, such as MGDA~\cite{DESIDERI2012313} and FAMO~\cite{liu2023famo}, solve the min-max problem in Equation~\ref{eq:weighted_loss} in an online fashion; further details are provided in Appendix~\ref{mtlgmm}. 
Therefore, these methods can be viewed as online approaches applicable to the data mixture problem.
Building on this insight, MFTCoder~\cite{liu2023mftcoderboostingcodellms} applies FAMO to multi-task instruction tuning for code LLMs, replacing the training loss with the validation loss for each domain when updating the domain weights $\weights^t$ across different domains.

\vspace{-0.1 cm}
Similarly, DoReMi~\cite{xie2023doremi} can also be operated in an online fashion by replacing the proxy models used for optimizing domain weights with the target model itself, enabling joint optimization of both domain weights and the target model throughout training. 
Consequently, some recent works~\cite{chen2024aioliunifiedoptimizationframework, belenki2025optimizingpretrainingdatamixtures} classify it as an online method.
ShearedLLaMa~\cite{xia2024shearedllamaacceleratinglanguage} improves upon this online version of DoReMi (referred to as DoReMi$_\text{online}$) by using the loss predicted by a fitted scaling law on each domain as reference loss and also replacing training loss with validation loss for each domain when updating $\weights^t$ during continued pretraining of LLMs.

\citet{luo2024velocitunevelocitybaseddynamicdomain} proposed Velocitune, a method that improves upon ShearedLLaMa.
They introduce the concept of learning velocity, defined as: $V^t_i =\frac{\ell_{\text{val}, i} \left(\modelparams^t \right)-\ell_{\text{ref}, i} }{\ell_{\text{init}, i}- \ell_{\text{ref}, i} }$, $i \in [k]$, to dynamically quantify how much the model has yet learned from each domain and then will be used to adjust the domain weights $\weights^t$ accordingly.
$\ell_{\text{init}, i}$ denotes the initial validation loss of the model $\modelparams$ on each domain $i$, while $\ell_{\text{ref}, i}$ is the reference loss, representing the final loss that the model is expected to achieve on each domain. 
As in ShearedLLaMa, the value of $\ell_{\text{ref}, i}$ is predicted using a scaling law on each domain, providing a cost-efficient estimation of the final loss.
We summarize different rules for updating $\weights^t$ across these methods in Table~\ref{online:min-max} to facilitate comparison; see Appendix~\ref{comparisons} for more details.

\vspace{-0.2cm}
\subsection{Online Mixing Law Methods}

\citet{chen2024aioliunifiedoptimizationframework} provide a unified perspective on several data mixture methods—including DML~\cite{ye2024datamixinglaws}, DoReMi$_\text{online}$, and Skill-It~\cite{chen2023skillitdatadrivenskillsframework}—by interpreting them as instances of optimizing a class of online mixing laws. 
These online mixing laws take the form of $\ell_{\text{val}, j}^{t+1}(\weights)=c_j^t+b_j^t \sigma\left(\sum_{i=1}^k -A_{i j}^t w_j^t\right), j \in[k]$, where the matrix $A^t \in \mathbb{R}^{k \times k}$ and the vectors $\boldsymbol{b}^t, \boldsymbol{c}^t \in \mathbb{R}^k$ are mixing law parameters specified by the respective methods.
The function $\sigma: \mathbb{R} \rightarrow \mathbb{R}$ is either the identity function or the exponential function.
By empirically examining their mixing law parameters with this unified formulation, \citet{chen2024aioliunifiedoptimizationframework} observe that these approaches specify the mixing law parameters inaccurately and result in poor performance.
To address this problem, they propose a simple online data mixing method \textsc{Aioli}, directly fitting linear mixing laws (i.e., $\sigma$ is the identity function) with the history of validation losses and domain weights during training: $\ell_{\text{val}, j}^{t+1}(\weights) = \ell_{\text{val}, j}^t(\weights) - \sum_{i=1}^k A_{ij}^t w_i^t, \quad \forall j \in [k]$ and uses the fitted laws to dynamically adjust domain weights $\weights^t$. We provide a comparison between Skill-It and Aioli in Table~\ref{online:mixing laws}.

\vspace{-0.2 cm}
\subsection{Others}
\vspace{-0.1 cm}

ODM~\cite{albalak2023efficientonlinedatamixing} and ADO~\cite{jiang2024adaptivedataoptimizationdynamic} adopt alternative strategies for dynamically adjusting domain weights during the training of the target model, differing from the two categories discussed earlier. 
Due to space constraints, we describe these methods in detail in Appendix~\ref{app:Others}.

\vspace{-0.2cm}
\subsection{Advantages and Disadvantages}
\vspace{-0.2cm}

Compared to offline methods, online methods do not suffer from scale-up issues.
When updating $\weights^t$, however, if the update rule involves computing model gradients, the computational overhead can be high. 
If it relies only on the model's loss, the overhead is generally lower, though it can still be significant depending on the specific implementation. This overhead can be mitigated by updating the weights intermittently—every $t_{\text{update}}$ steps, where $t_{\text{update}} > 1$—instead of every step.
Another limitation is that when out-of-domain loss is used to evaluate model performance, only online mixing law methods are applicable, while most other online methods (e.g., online min-max optimization methods) cannot be used in this setting.

\vspace{-0.2cm}
\section{Challenges and Future Directions}

\vspace{-0.2cm}
\subsection{Difficulty in finding the optimal domain weights}
\vspace{-0.1cm}

Offline methods assume that there exist optimal static domain weights $\bestweights$ for a given corpus $\mathcal D_{\text{train}}$ composed of different domains that lead to the best-performing model.
\citet{jiang2024adaptivedataoptimizationdynamic} argue that the optimal domain weights $\bestweights$ on $\mathcal D_{\text{train}}$ form a fixed, stationary curriculum throughout the training process, where models are progressively exposed to each domain in a curated order, which is closely related to curriculum learning~\cite{bengio2009curriculum} and empirically demonstrate the existence of effective training curricula. 
However, identifying the globally optimal curriculum is computationally challenging, making it highly unlikely to be found. 
This raises the question of how to identify a good curriculum in a computationally efficient manner. 
In response, \citet{jiang2024adaptivedataoptimizationdynamic} suggest adopting an online strategy for optimizing domain weights and training large language models, which could effectively identify good curricula while minimizing computational costs.

\vspace{-0.2cm}
\subsection{The natural domain}
% \vspace{-0.2cm}

Most existing works optimize domain weights to the natural domains of the data, which could bring two issues:
(1) The natural domains are typically at a coarse-grained level, which should be divided into fine-grained domains to allow for gains from applying various algorithms. 
For example, the GLaM dataset~\cite{du2022glamefficientscalinglanguage} has 8 domains, while the Pile~\cite{gao2020pile800gbdatasetdiverse} is constructed from 22 diverse domains. 
The DoReMi algorithm~\cite{xie2023doremi} performs better on the Pile than on the GLaM dataset.
(2) How can we find fine-grained domains in the training data? One direct way is through clustering~\cite{oren-etal-2019-distributionally, gururangan2023scalingexpertlanguagemodels, chen2023skillitdatadrivenskillsframework}.
In \citet{gururangan2023scalingexpertlanguagemodels}, the authors split a corpus into segments and use the embeddings of these segments as sample features to run $k$-means clustering algorithms for unsupervised domain discovery.
\citet{chen2023skillitdatadrivenskillsframework} explore several sample features for clustering the data, including the embedding of the last token, the average token, the last token of the training sample, and the loss trajectory of the training sample over multiple runs. They find that using the loss trajectory as the sample feature significantly outperforms the other metrics.

\vspace{-0.2cm}
\subsection{Loss metrics vs. Task performance}
% \vspace{-0.2cm}

To determine the optimal domain weights, loss metrics, such as training loss or validation loss, are often used as surrogate objectives for the downstream task performance to evaluate the model during training.
However, does the model trained with the optimal domain weights found by evaluating the loss really perform well on the downstream tasks?
\citet{gadre2024languagemodelsscalereliably} explore this issue within the framework of scaling laws. 
They find that the perplexity of a language model can be related to its downstream task performance via a power law, demonstrating that it is reasonable to use loss metrics as surrogate objectives for downstream task performance when determining the best training variables, such as model size or dataset size.
However, the optimal domain weights found by most methods in data mixture are only slightly better than uniform sampling~\cite{jiang2024adaptivedataoptimizationdynamic}, or even worse than uniform sampling~\cite{chen2024aioliunifiedoptimizationframework}.
We think it is worth exploring the relationship between loss metrics and task performance under the data mixture setting.

\vspace{-0.1cm}
\section{Conclusion}
\vspace{-0.1cm}

We provide an overview of existing data mixture methods from an algorithm perspective in this paper.
We first propose a fine-grained categorization of existing methods, extending beyond the previous offline/online classification.
We divide offline methods into heuristic-based, algorithm-based, and function fitting-based methods. 
We categorize online methods into three groups—online min-max optimization, online mixing law, and other approaches—by drawing connections with the optimization frameworks underlying offline methods.
We then summarize the problem formulations, representative algorithms for each subtype of offline and online methods, and clarify the relationships and distinctions among them.
Finally, we discuss the advantages and disadvantages of each method and highlight key challenges in data mixture.

\newpage

\section*{Limitation}
We only provide insights into data mixture problems from theoretical perspectives and do not conduct any empirical studies. 
Most methods emphasize their own strengths, but their comparisons are not aligned in terms of experimental settings, such as model type, model scale, and training data scale. 
Even though we discuss the advantages and disadvantages of each method, we have not sufficiently compared them from an experimental perspective.

\bibliography{custom}

\newpage

\appendix

% \onecolumn

\section{Static Weights vs. Dynamic Weights}
\label{static-dynamic}

\paragraph{Static weights.} 
Static weights are primarily used to combine the training data $\mathcal D_{\text{train}}$ by assigning fixed sampling probabilities to each domain during model training. 
These weights remain unchanged throughout the entire training process.  
In offline methods, the optimal domain weights $\bestweights$ used to train the target model are considered static weights.

\paragraph{Dynamic weights.} 
Dynamic weights are adaptively updated during model training, in contrast to static weights that remain fixed.
However, directly adjusting the data mixing ratio for all domains within each mini-batch during training is not straightforward.
A more common approach is to apply uniform sampling to sample all domains in $\mathcal D_{\text{train}}$ and alter the data contribution through dynamic weighting of the loss function, which can be expressed as:
$\sum_{i=1}^{k} w_i^t \ell_{\text{train}, i} (\modelparams^t)$. 
Here $\ell_{\text{train}, i} (\modelparams^t)$ represents the average loss of $\modelparams^t$ on the $i$-th domain of the mini-batch data at each training step $t$. 
The domain weights $\weights^t$ used in online methods are typically dynamic weights, which will be updated at each training iteration or a fixed interval $t_{\text{update}}$.

\section{Offline Methods}

\subsection{Heuristic Methods}
\label{app:Heuristic}

Heuristic-based methods, such as uniform sampling and proportional sampling, determine domain weights based on simple heuristics.

Uniform sampling defines the optimal static domain weights $\bestweights=[\frac{1}{k}, \dots, \frac{1}{k}]$ when mixing $k$ training domains $\mathcal D_{\text{train}} = \left\{D_{\text{train}, 1}, \ldots, D_{\text{train}, k}\right\}$. 
Proportional sampling is commonly used in large-scale training, where domain weights are set in proportion to token counts. Specifically, the best domain weights $\bestweights$ are defined as $w_i^{\star} = \frac{|D_{\text{train}, i}|}{\sum_{j=1}^k |D_{\text{train}, j}|}$, $i \in [k]$, where $|D_{\text{train}, i}|$ denotes the number of tokens in each training domain data $D_{\text{train}, i}$.

The domain weights in these methods remain fixed rather than being optimized and thus they do not rely on proxy models. Considering that they can be implemented by omitting line 3 from Framework~\ref{offline}, and for the sake of completeness in our framework, we treat them as degenerate cases of offline methods.

\subsection{Algorithm-based Methods}

\subsubsection{Bi-level Optimization}
\label{app:bi-lev}

\begin{equation}
% \label{eq:bi-level}
\begin{aligned}
& \boldsymbol{w}^{\star} \in \underset{\boldsymbol{w} \in \Delta^k}{\arg \min } \ell_{\text{val}} \left(\proxyparams^{\star}(\boldsymbol{w})\right) \\
& \quad \text { s.t. } \proxyparams^{\star}(\boldsymbol{w}) \in \underset{\proxyparams}{\arg \min } \sum_{i=1}^k w_{i} \ell_{\text{train}, i}(\proxyparams).
\end{aligned}
\end{equation}
There are three primary optimization frameworks~\cite{zhang2023introductionbileveloptimizationfoundations} used to solve the bi-level optimization problems: the Implicit Function (IF)-based approach, the Gradient Unrolling (GU)-based approach, and the Value Function (VF)-based approach.
The primary obstacle hindering the scalability of bi-level optimization stems from the interdependence between the outer-level and inner-level problems. This mutual dependency introduces significant computational challenges, such as the need to compute the Hessian and Jacobian, especially when handling large-scale problem instances.
The first two classes both leverage (some approximated version of) the Implicit Gradient (IG).
The key difference is how the IG approximation is conducted: One directly assumes that a given procedure can provide a high-quality solution to the inner-level problem, while the other approximates the inner-level solution by unrolling a given algorithm for a fixed number of steps. 
The Value Function (VF)-based approach reformulates the bi-level optimization as a single-level regularized optimization problem, which offers flexibility in handling inner-level constraints.

\subsection{Function Fitting-based Methods}
\label{app:mixing-laws}

\subsubsection{The D-CPT law}
\label{CD-CPT-law}

\citet{que2024dcptlawdomainspecificcontinual} study mixing laws for Domain-Specific Continual Pre-Training (D-CPT) by following the Chinchilla scaling law~\cite{hoffmann2022trainingcomputeoptimallargelanguage}. In the simplest scenario, where the training data consists of a general corpus $\mathcal{D}_{\text{train}, 0}$ and a domain-specific corpus $\mathcal{D}_{\text{train}, 1}$, they model the mixing law $f_j, j\in \{ 0, 1 \}$ as:
\begin{equation}
\ell_{\text{val}, j}(w_j, D, N)=E^j+\frac{A^j}{N^{\alpha^j}}+\frac{B^j \cdot w_j^\eta}{D^{\beta^j}}+\frac{C^j}{w_j^{\prime}}, 
\end{equation}
where $D$ is dataset size, $N$ is model size, $w_j^{\prime} = w_j + \epsilon$, and $\epsilon$ is a small constant used to ensure the stability of $\ell_{\text{val}, j}(N, D, w_j)$ when $w_j$ approaches zero.
The set of parameters $\boldsymbol{E}, \boldsymbol{A}, \boldsymbol{B}, \boldsymbol{C}, \boldsymbol{\alpha}, \boldsymbol{\beta}, \boldsymbol{\eta} \in \mathbb{R}^2$ is learned by fitting the model using the L-BFGS optimization algorithm~\cite{liu1989limited}.

They propose three designs for the aggregation function $g$ to obtain the optimal static domain weights $\bestweights$. 
Among them, the most widely adopted strategy typically considers a trade-off between the general and domain-specific abilities of the model after Continued Pre-Training. 
In this approach, the optimal proportion for the domain corpus is determined as $w^{\star}_1 = \underset{w_1}{\operatorname{argmin}} \ell_{\text{val}, 1} \left(N=N_0, D=D_0, w_1 \right)$ $\text { s.t. } \frac{\ell_{\text{val}, 0}-\ell_{\text{val}, 0}^0}{\ell_{\text{val}, 0}^0}<T$, where $T$ denotes a threshold that constrains the degradation in the model's general ability. 
Therefore, the optimal domain weights is $\bestweights = (1- w^{\star}_1, w^{\star}_1)$.

\section{Online Methods}

\subsection{Online Min-Max Optimization Methods}

\subsubsection{Gradient Manipulation Methods}
\label{mtlgmm}

A class of methods in multi-task learning, known as gradient manipulation methods, solves Equation~\ref{eq:weighted_loss_variant} by finding a new update $d_t$ at each step in an online fashion to prevent task optimization unbalance:
\begin{equation}
\label{eq:weighted_loss_variant}
\begin{aligned}
\min _{\theta} \max_{\boldsymbol{w}^t \in \Delta^k} \sum_{i=1}^k w^t_i f^t_i(\modelparams),
\end{aligned}
\end{equation}
where $f_i^t(\modelparams)$ is related with $\ell^t_{\text{train},i}(\modelparams)$. 
In a gradient descent style iterative update: $\modelparams^{t+1}=\modelparams^t-\eta d_t$, $\eta$ is the learning rate, $d_t = \sum_i w^t_i \nabla \ell_{\text{train}, i}^t(\modelparams)$.
There are several approaches~\cite{DESIDERI2012313, liu2021conflict, liu2021towards} targeting to improve the "worst-case improvement" among all tasks, where $f_i^t(\modelparams)= \ell_{\text{train}, i}^t(\modelparams) -\ell_{\text{train}, i+1}^t(\modelparams)$. 
\citet{yu2020gradient} propose a form of gradient surgery to avoid detrimental gradient interference, also helping improve the "worst improvement" across all tasks.
\citet{liu2023famo} propose a method called FAMO to optimize the "largest worst-case improvement rate" across all tasks to achieve better task balance, $f^t_i(\modelparams)= (\ell_{\text{train}, i}^t(\modelparams) -\ell_{\text{train}, i+1}^t(\modelparams)) / \ell_{\text{train}, i}^t(\modelparams)$.

\subsubsection{Comparison of Online Min-Max Optimization Methods}
\label{comparisons}

DoReMi$_\text{online}$ jointly optimizes both the target model $\modelparams$ and the domain weights $\weights$ during training.
Following the framework of Algorithm~\ref{online}, the target model $\modelparams$ is updated as 
\begin{equation}
\modelparams^{t} = \modelparams^{t-1} - \eta \sum_{i=1}^k w_i^{t-1} \nabla \ell_{\text{train}, i}(\modelparams^{t-1}), \nonumber
\end{equation}
while the dynamic domain weights $\weights^t$ are updated according to 
\begin{equation}
\small
\begin{aligned}
\weights^t & = \frac{\boldsymbol{\alpha}^t}{\sum_{i=1}^k \alpha^t_i}, \text{ where } \boldsymbol{\alpha}^t =  \\ 
& \boldsymbol{\alpha}^{t-1} \exp{ (\zeta \max \{ \ell_{\text{train}, i} \left(\modelparams^t \right)-\ell_{\text{train}, i}(\modelparams_{\text{ref}}), 0 \}) },
\end{aligned} \nonumber
\end{equation}
where $\zeta$ is a step size, and $\modelparams_{\text{ref}}$ is a pre-trained reference model, typically of the same size as the target model $\modelparams$.
The reference model's training loss $\ell_{\text{train}, i}(\modelparams_{\text{ref}})$ serves as a baseline for evaluating the target model’s current performance. 
More specifically, the excess loss $\ell_{\text{train}, i}(\modelparams^t) - \ell_{\text{train}, i}(\modelparams_{\text{ref}}) $ measures how much the target model has yet to learn from domain $i$.
For ease of comparison with Velocitune~\cite{luo2024velocitunevelocitybaseddynamicdomain}, we also denote this quantity as: $V^t_i = \ell_{\text{train}, i}(\modelparams^t) - \ell_{\text{train}, i}(\modelparams_{\text{ref}})$.

Training a reference model $\modelparams_{\text{ref}}$ to obtain baseline losses for evaluating the learning progress of the target model on each domain can be prohibitively expensive.
To reduce this cost, ShearedLLaMa~\cite{xia2024shearedllamaacceleratinglanguage} improves upon DoReMi$_\text{online}$ by using losses predicted by a fitted scaling law for each domain as the reference losses.
Additionally, it replaces the training loss $\ell_{\text{train}, i}(\modelparams^t)$ with the validation loss $\ell_{\text{val}, i}(\modelparams^t)$ when dynamically updating the domain weights $\weights^t$ to optimize the generalization ability of the target model (i.e., $V^t_i = \ell_{\text{val}, i}(\modelparams^t) - \ell_{\text{ref}, i}$).

Velocitune~\cite{luo2024velocitunevelocitybaseddynamicdomain} introduces a more precise metric to measure how much the target model has yet to learn from each domain, defined as $V^t_i =\frac{\ell_{\text{val}, i} \left(\modelparams^t \right)-\ell_{\text{ref}, i} }{\ell_{\text{init}, i}- \ell_{\text{ref}, i} }, i \in [k]$, where $\ell_{\text{init}, i}$ denotes the initial validation loss for domain $i$, and $\ell_{\text{ref}, i}$ denotes the reference loss. 
Similar to ShearedLLaMa, $\ell_{\text{ref}, i}$ is estimated using a fitted scaling law on each domain $i$, providing a cost-efficient approximation of the desired learning target for each domain.

\subsection{Others}
\label{app:Others}

\subsubsection{Online Data Mixing (ODM)}
\label{app.odm}

\citet{albalak2023efficientonlinedatamixing} propose an efficient method, Online Data Mixing (ODM), which leverages multi-armed bandit (MAB) algorithms by treating each data domain as an arm.
In ODM, the principle of updating $\weights^t$ at each training step $t$ is to increase the mixing ratio for domains that provide the most valuable information for model training.
Drawing from information theory that perplexity can be interpreted as the expected information gain from learning the next token, they use the training loss per domain as a reward signal. 
The ODM method is ultimately modified based on the Exponential-weight algorithm for Exploration and Exploitation (Exp3)~\cite{auer2002nonstochastic}, with the key difference being that ODM employs a moving average estimated reward instead of a cumulative estimated reward, paying more attention to the recent rewards.

\subsubsection{Adaptive Data Optimization (ADO)}
\label{app.ado}

\citet{jiang2024adaptivedataoptimizationdynamic} propose Adaptive Data Optimization (ADO), where they adjust the dynamic weights $\weights^t$ at a fixed interval $t_{\text{update}}$ by fitting a domain scaling law for each domain.
The domain scaling law predicts the $i$-th domain's training loss after training on $n$ samples: $\widehat{\ell_{\text{train}, i}} (n)=\widehat{\ell}\left(n; \alpha_i, \beta_i, \varepsilon_i\right)=$ $\varepsilon_i + \beta_i n^{-\alpha_i}$, $\varepsilon_i$ corresponds to the estimated irreducible loss of domain $i$.
The derivative of the loss to the number of samples can be interpreted as follows:
\begin{equation}
\begin{aligned}
\frac{d \widehat{\ell_{\text{train}, i}} (n)}{d n}&=\frac{-\alpha_i \beta_i n^{-\alpha_i}}{n} \\
&=-\frac{1}{n} \underbrace{\alpha_i}_{\text{Learning speed}} \underbrace{\left(\widehat{\ell_{\text{train}, i}} (n)-\varepsilon_i\right)}_{\text{Reducible loss}} \nonumber
\end{aligned}
\end{equation}
To determine $\weights^t$ during model training, they define a preference distribution
\begin{equation}
\rho_i^t \propto -\mu_i \frac{\partial}{\partial n} \widehat{\ell_{\text{train}, i}} (n) \lambda_i^t ,
\end{equation}
here $\mu_i$ is the prior domain weights, and $\frac{\partial}{\partial n} \widehat{\ell_{\text{train}, i}} (n)$ is the learning speed forecast by a scaling law.
$\lambda_i^t$ is a credit assignment score for domain $i$, which is a real positive-valued function that indicates how much data from the $i$-th domain contributed to recent changes in the loss $\mathcal{L}_i$.
The intuition behind ADO is to prioritize sampling from a domain $i$ if the model is decreasing its loss $\mathcal{L}_i$ quickly, but only if that decrease can be attributed to data from domain $i$.
The dynamic weights $\weights^t$ is a linear combination of the moving average of preference policy $\rho^t$ at every step.

\section{Differences between the previous survey paper and our paper}
\label{survey-diff}

\citet{albalak2024surveydataselectionlanguage} is a comprehensive survey on data selection, with data mixture being a small part of it.
They categorize existing data mixture methods into two main types: offline methods and online methods.
For each category, they discuss a limited number of methods, mainly because there were not many related works available at that time.

For our survey paper, we propose a fine-grained categorization of existing data mixture methods, extending beyond their offline and online classification to include and contextualize a broader range of recent and subsequent developments.
Specifically, we divide offline methods into three subcategories: heuristic-based, algorithm-based, and function fitting-based methods.
For algorithm-based methods, we systematically analyze their connections to min-max optimization and bi-level optimization. For function fitting-based methods, we identify three key components and present a formal problem formulation.
We further compare min-max optimization methods, bi-level optimization methods, and function fitting-based approaches, and highlight their interrelationships. In particular, we point out that the key distinction between min-max and bi-level optimization lies in how the optimal domain weights are defined, while function fitting-based methods convert the bi-level optimization problem into a single-level problem via function fitting.

We further draw connections between online methods and the optimization frameworks underlying offline methods by interpreting most online approaches as online variants of offline techniques. Based on this perspective, we categorize online methods into three groups: online min-max optimization, online mixing law, and other methods.
We also provide a systematic comparison and summary of online min-max optimization methods and online mixing law methods, highlighting their differences in objective design, update strategies.

Finally, we survey these methods primarily from an algorithmic perspective, aiming to provide deeper understanding and valuable insights into optimization strategies for data mixture, as we believe they are applicable to any stage of model training.

\end{document}